\documentclass[10pt,twocolumn,letterpaper]{article}

\usepackage{wacv}
\usepackage{times}
\usepackage{epsfig}
\usepackage{graphicx}
\usepackage{amsmath}
\usepackage{amssymb}

\usepackage{kotex}
\usepackage{booktabs}
\usepackage{tabularx}
\usepackage{multirow}
\usepackage{arydshln}
\usepackage{algorithm}
\usepackage{algpseudocode}
\usepackage{bm}
\usepackage{subcaption}
\usepackage{stfloats}
\usepackage{arydshln}

\DeclareMathAlphabet      {\mathsb}{OT1}{cmr}{b}{n}
\newcommand{\vcsb}[1]{\ensuremath{\mathsb{#1}}} 
\DeclareMathOperator*{\argmax}{argmax}

\newcommand*{\boldv}{\ensuremath{\bm{v}}}
\newcommand*{\boldw}{\ensuremath{\bm{w}}}
\newcommand*{\boldx}{\ensuremath{\bm{x}}}
\newcommand*{\boldz}{\ensuremath{\bm{z}}}
\newcommand*{\boldSigma}{\ensuremath{\bm{\Sigma}}}
\newcommand*{\boldtheta}{\ensuremath{\bm{\theta}}}
\newcommand*{\boldomega}{\ensuremath{\bm{\omega}}}
\newcommand*{\boldpsi}{\ensuremath{\bm{\psi}}}
\newcommand*{\boldphi}{\ensuremath{\bm{\phi}}}
\newcommand*{\boldnu}{\ensuremath{\bm{\nu}}}
\newcommand*{\task}{\ensuremath{\mathcal{T}_i}}
\newcommand*{\support}{\ensuremath{S_{i}}}
\newcommand*{\query}{\ensuremath{Q_{i}}}

\def\wacvPaperID{0317} 

\wacvfinalcopy 

\ifwacvfinal
\fi

\ifwacvfinal
\usepackage[breaklinks=true,bookmarks=false]{hyperref}
\else
\usepackage[pagebackref=true,breaklinks=true,colorlinks,bookmarks=false]{hyperref}
\fi

\begin{document}
\title{Contextual Gradient Scaling for Few-Shot Learning}

\author{Sanghyuk Lee \qquad\quad Seunghyun Lee \qquad\quad Byung Cheol Song \\
Department of Electrical and Computer Engineering, Inha University \\
{\tt\small \{sanghyuk.lee625, lsh910703\}@gmail.com, bcsong@inha.ac.kr}}

\maketitle

\ifwacvfinal
\thispagestyle{empty}
\fi

\begin{abstract}
    Model-agnostic meta-learning (MAML) is a well-known optimization-based meta-learning algorithm that works well in various computer vision tasks, e.g., few-shot classification. MAML is to learn an initialization so that a model can adapt to a new task in a few steps. However, since the gradient norm of a classifier (head) is much bigger than those of backbone layers, the model focuses on learning the decision boundary of the classifier with similar representations. Furthermore, gradient norms of high-level layers are small than those of the other layers. So, the backbone of MAML usually learns task-generic features, which results in deteriorated adaptation performance in the inner-loop. To resolve or mitigate this problem, we propose contextual gradient scaling (CxGrad), which scales gradient norms of the backbone to facilitate learning task-specific knowledge in the inner-loop. Since the scaling factors are generated from task-conditioned parameters, gradient norms of the backbone can be scaled in a task-wise fashion. Experimental results show that CxGrad effectively encourages the backbone to learn task-specific knowledge in the inner-loop and improves the performance of MAML up to a significant margin in both same- and cross-domain few-shot classification. 
\end{abstract}

\section{Introduction}
\label{sec:intro}

\begin{figure}
\begin{center}
    \begin{subfigure}[t]{\linewidth}
        \centering
        \includegraphics[width=.9\linewidth]{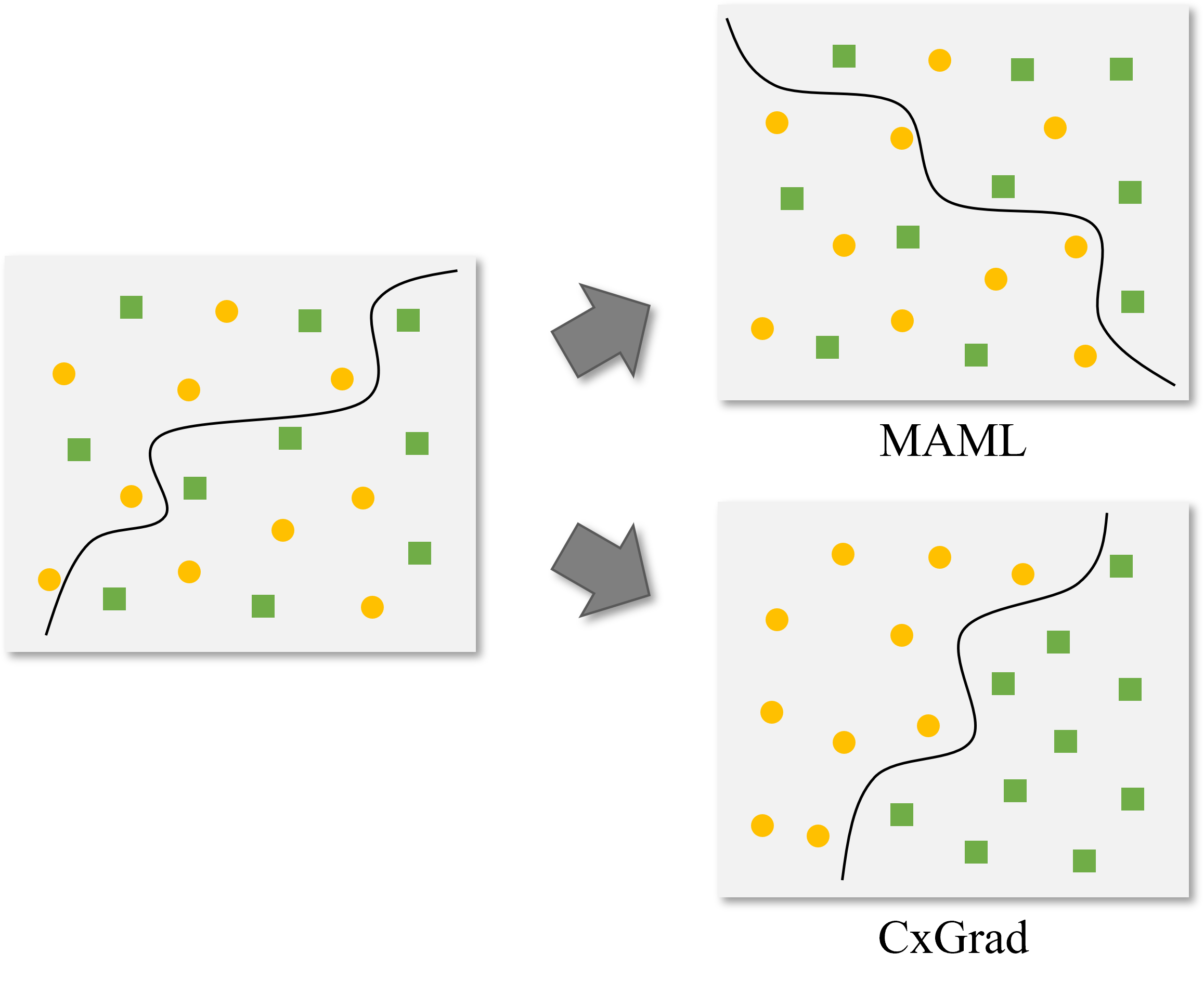}
        \setlength{\abovecaptionskip}{2pt}
        \setlength{\belowcaptionskip}{10pt}
    \end{subfigure}
\end{center}
    \setlength{\abovecaptionskip}{0pt}
    \caption{An overview of the adaptation scheme of MAML and CxGrad. Shapes represent samples belonging to different classes, and lines represent decision boundaries of the classifier. MAML focuses on updating the decision boundary to adapt to a given task. However, CxGrad mainly updates the feature extractor rather than the decision boundary. As a result, although training tasks and target tasks are far different, we can get optimized representation for the target tasks. Furthermore, CxGrad updates the decision boundary based on the changed representation.}
\label{fig:concept}
\end{figure}
    Deep learning has enabled rapid development of various computer vision tasks such as classification~\cite{resnet,alexnet,EfficientNet}, object detection~\cite{RCNN,ssd,yolo}, and segmentation~\cite{mask-RCNN,FCN,U-net}. However, the success of many deep learning-based computer vision algorithms highly counts on large datasets. Unfortunately, there is no guarantee that we can always collect large-scale data due to privacy issues or the low frequency of occurrence of the target event.
    
    The critical difference between machines and humans in the learning process is whether or not to use prior knowledge. Machines are not good at leveraging prior knowledge (learned in the past) on current learning tasks, while humans can learn new tasks quickly by utilizing prior knowledge. 
    
    To bridge this gap, few-shot learning has been developed~\cite{FSL_survey}. Especially, meta-learning draws a lot of attention. Meta-learning learns prior knowledge, also known as meta-knowledge~\cite{survey}, through various tasks. That is, transferring prior knowledge to a new task enables more efficient learning of the task. As a famous meta-learning, MAML~\cite{MAML} learns an initialization of a model as meta-knowledge. Thus, MAML makes the adaptation to a new task complete in only a few steps.
    
    However, MAML does not achieve effective adaptation due to the following two problems related to gradient norms. The first problem is that as meta-training progresses, the gradient norm of the classifier increases while those of the backbone decrease. Note that a gradient norm implies to what extent a certain layer has to be updated. As the gradient norm of the classifier becomes larger than the backbone, the representation reuse and the decision boundary-oriented update tend to dominate further and further. As a result, since good representations are unavailable and the adaptation highly depends on the decision boundary during learning, a new task cannot be solved properly. In short, this problem makes effective adaptation difficult. The second problem is that gradient norms of high-level layers of the backbone are much smaller than those of the low- and mid-level layers. It is well known that high-level layers are good at learning discriminative features~\cite{high-level_layer}. Thus, this problem indicates that the backbone focuses on extracting generic and low-level features rather than discriminative and high-level features from a given task. 
     
    Due to the problems mentioned above, MAML is tempted to learn task-generic representations and its adaption strongly relies on the decision boundary-oriented update. So it is likely to perform unsatisfactory adaptation in both meta-training and meta-testing. This phenomenon becomes severer when dataset domains in meta-training and meta-testing are different from each other (e.g., cross-domain). Meanwhile, note that our argument is closely related to what is said in \cite{BOIL}. \cite{BOIL} insists that an appropriate representation change is required for MAML. Actually, BOIL~\cite{BOIL} froze the classifier during adaptation to change representations as much as possible. However, BOIL still has a limit to leverage the representation change because it does not explicitly handle gradient norms.
    
    To solve the problems mentioned above, this paper proposes Contextual Gradient Scaling (CxGrad). CxGrad learns the task-embedding vectors by adopting context parameters~\cite{CAVIA} in adaptation. Specifically, a sub-network takes context parameters as input and generates scaling factors to scale gradient norms of the backbone in a task-wise manner. Since the sub-network consists of differentiable layers and allows end-to-end learning, CxGrad can find the local optimal scaling factors to increase the proportion of the backbone in meta-training. Figure~\ref{fig:concept} compares how MAML and CxGrad perform adaptation in the inner-loop, respectively. We can observe that CxGrad facilitates backbone learning to get a more optimized representation than MAML. Note that the dispersed samples make clusters after adaptation. Therefore, CxGrad makes the backbone learn task-specific features, which results in the performance improvement of MAML in both same- and cross-domain few-shot classification.
    
    The contribution points of this paper are summarized as follows:
    \begin{itemize}
        \item We propose a meta-learning algorithm that contextually scales gradient norms. The proposed scaling method decreases gradient norms of the classifier and increases those of the backbone for effective adaptation.
        \item Furthermore, the proposed method learns more task-specific features than MAML by increasing gradient norms of high-level layers of the backbone.
        \item As a result, the proposed method provides state-of-the-art or very competitive performance in both same- and cross-domain few-shot classification.
    \end{itemize}

\section{Related Work}
\label{sec:related}
    The goal of meta-learning, also known as learning to learn~\cite{learning_to_learn}, is to utilize meta-knowledge shared between tasks by transferring it to a new task for more efficient adaptation~\cite{bengio1,bengio2,schumidhuber2,schmudhuber3,learning_to_learn}. In general, meta-learning can be classified into three categories: metric-based, model-based, and optimization-based algorithms. Metric-based algorithms prefer adaptation through an embedding function that maps each input into an embedding space~\cite{siamese,ProtoNet,RelationNet,MatchingNet}. The embedding function learns to make similar classes mapped close to each other and different classes mapped far from each other. Specifically, it embeds the support set and the query set, and then compares their similarities. Model-based algorithms encode the adaptation process into the feed-forward path. They parameterize a task by encoding the training set~\cite{meta-networks}, or use a separate buffer~\cite{MANN}. Optimization-based algorithms literally optimize the adaptation process~\cite{L2L_by_GD,MAML,optim_as_model}.
    
    The most representative optimization-based method is MAML~\cite{MAML}. Since MAML can be applied to various computer vision fields based on optimization, it has recently received considerable attention, along with few-shot classification. MAML learns an initialization as meta-knowledge that can be adapted to a new task in a small number of steps. Still, MAML suffers from a few drawbacks, such as requiring high computation cost due to bi-level optimization and lower performance than other meta-learning methods.
    
    This study pursues task-dependent meta-learning. Zintgraf~\cite{CAVIA} proposes context parameters, i.e., a low-dimensional task representation, which embed task information in a vector form. Context parameters are additionally fed into a given model to generate factors for modulating the intermediate feature maps. ModGrad~\cite{ModGrad} adopts context parameters of \cite{CAVIA} to solve the noisy gradient problem in the low-data regime by element-wise gradient modulation. \cite{L2F} observes that the direction of updating meta-knowledge between tasks causes conflicts and proposes a method called L2F to attenuate the influence of meta-knowledge in a task-dependent manner. In detail, L2F generates scaling factors and then scales the model parameters with them.
    
    Also, this study is closely related to algorithms using auxiliary task-dependent information during backpropagation. For example, Meta-SGD~\cite{meta-sgd} chooses learning rates and update directions as meta-knowledge and achieves better generalization ability than MAML. However, since each parameter is updated with its own learning rate, the amount of parameters to be learned becomes doubled. MAML++~\cite{MAML++} focuses on the instability issue that MAML suffers in meta-training and then presents a solution using multi-step loss optimization, cosine annealing of learning rate~\cite{SGDR}, and so on. Also, since MAML++ learns the learning rate for each layer and each step, it can reduce the cost of parameter size, resulting in overcoming the disadvantage of Meta-SGD.
    
    Besides, we try to change the representation for each task. Recently, two studies~\cite{BOIL,ANIL} made contradictory claims about the necessity of the inner-loop in MAML. \cite{ANIL} showed that MAML could be sufficiently learned without inner-loop by reusing features. However, ~\cite{ANIL} didn't provide any experimental result regarding cross-domain. On the other hand, \cite{BOIL} induced representation change in the inner-loop by freezing the classifier and showed that the inner-loop plays an important role in solving cross-domain few-shot classification. 

    The proposed method can be differentiated from the methods mentioned above in some respects. First, in order to induce task-specific feature learning, L2F~\cite{L2F} attenuates conflicts among tasks, but the proposed method scales gradient norms of the backbone by using the property of batch normalization (BN)~\cite{batchnorm}. Second, while CAVIA~\cite{CAVIA} and ModGrad~\cite{ModGrad} update context parameters in a single step, the proposed method makes the context parameters depend on all steps by accumulating the information obtained from the meta-learner at each step. The third one is the difference in the representation change perspective. Although BOIL~\cite{BOIL} emphasizes the importance of the representation change for the first time, its representation change is very limited only by freezing the classifier. On the contrary, assuming that a major factor interfering with representation change results from gradient norms, the proposed method generates task-wise scaling factors in an end-to-end manner and then scales gradient norms of the backbone. Thus, more effective representation change can be derived.
    
\section{Proposed Method}
\label{sec3:method}
    \begin{figure*}[ht]
\begin{center}
    \begin{subfigure}[t]{.45\linewidth}
        \centering
        \includegraphics[width=1\linewidth]{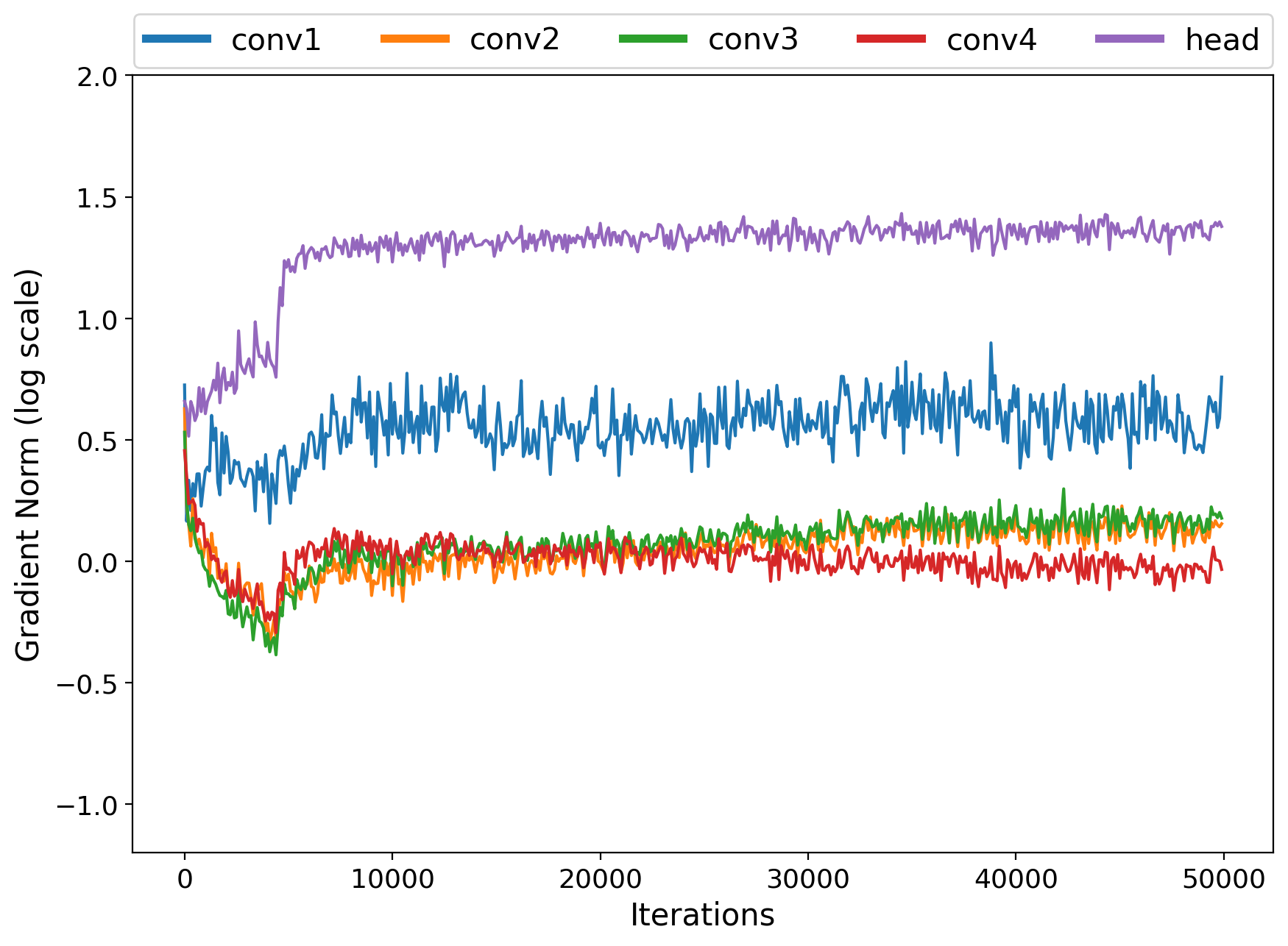}
        \caption{MAML}
        \label{fig:gradient_norm_MAML}
    \end{subfigure}
    \begin{subfigure}[t]{.45\linewidth}
        \centering
        \includegraphics[width=1\linewidth]{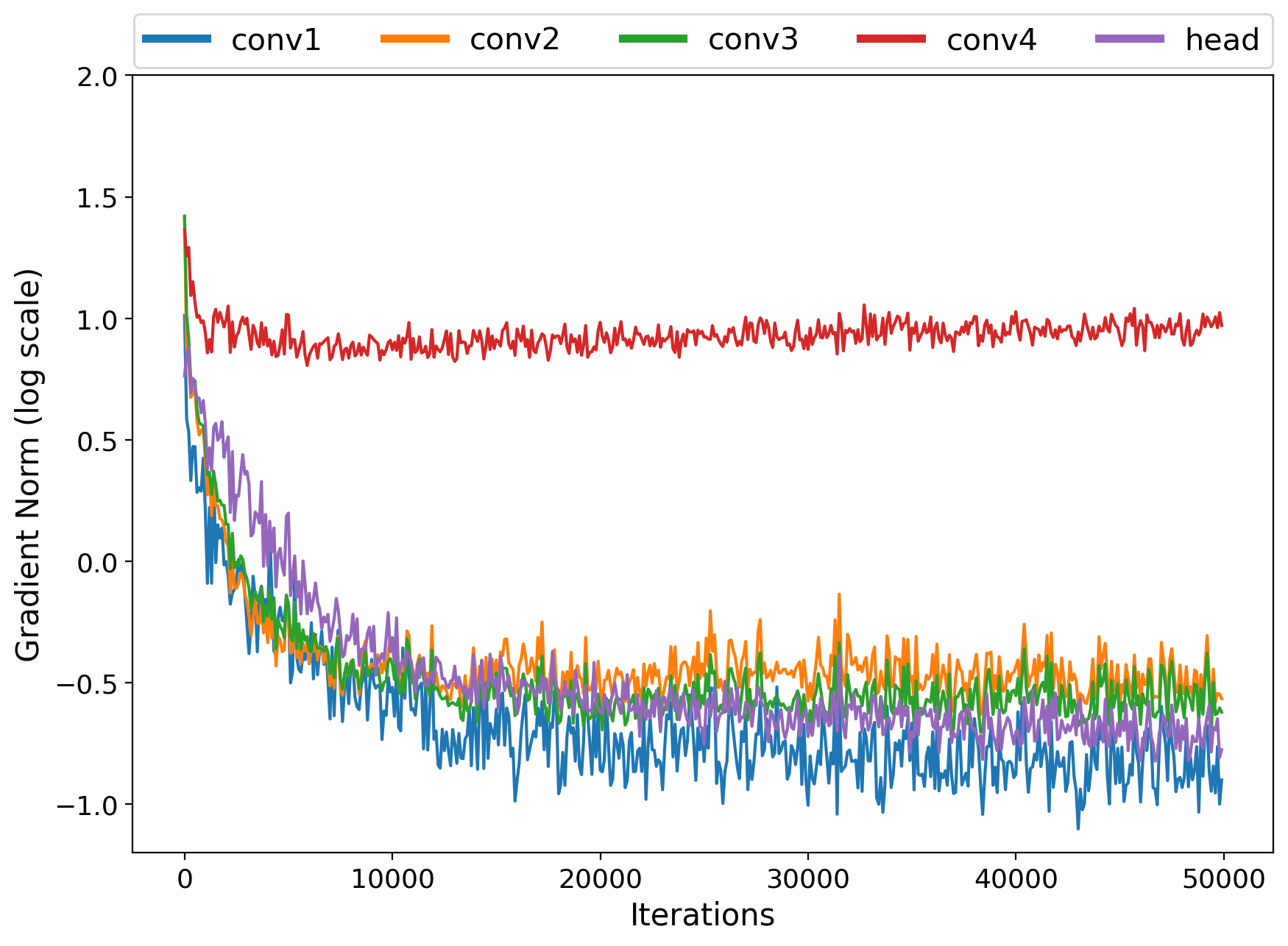}
        \caption{CxGrad}
        \label{fig:gradient_norm_CxGrad}
    \end{subfigure}
\end{center}
    \setlength{\abovecaptionskip}{0pt}
    \caption{Gradient norms of MAML and CxGrad during adaptation in meta-training. Gradient norms are averaged over inner-loop steps and tasks of the batch. In (a) MAML, the gradient norm of the classifier is bigger than those of backbone layers. Furthermore, gradient norms of low- and mid-levels are bigger than those of high-level layers. Considering the fact that high-level layers learn more discriminative features, MAML reuses similar representation from meta-knowledge and heavily depends on updating a decision boundary. However, in (b) CxGrad, gradient norms of high-level layers are much bigger than the others. Also, the norm gets smaller near the front side of the backbone. As a result, CxGrad focuses on changing representation and the backbone learns to extract task-specific features.}
\label{fig:gradient_norm}
\end{figure*}
    \subsection{Preliminaries}
    \label{sec3.1:preliminaries}
        Before explaining the details of the proposed method, we describe overviews of meta-learning and MAML in this section. Suppose there exists a distribution over tasks $p(\mathcal{T})$. The goal of meta-learning is to learn transferable meta-knowledge from $p(\mathcal{T})$. For this, we compose a mini-batch $\{ \task \}_{i=1}^{B}$ with multiple tasks sampled from $p(\mathcal{T})$ and $B$ is the batch size. Here, $i$ is the index of a task. Each $\task$ consists of a task-specific loss function $\mathcal{L}_{\task}$, a support set $\support$, and a query set $\query$:

        \begin{equation}
            \support = \{ (\bm{x}_{j}, y_{j} ) \}_{j=1}^{|\support|}, \quad
            \query = \{ (\bm{x}_{j}, y_{j}) \}_{j=1}^{|\query|}
        \end{equation}
        
        \noindent where $|\cdot|$ is the cardinality of a set, $\bm{x}_{j} \in \mathcal{X}$ is a sample of the input space $\mathcal{X}$, and $y_{j} \in \mathcal{Y}$ is the corresponding label. $\mathcal{Y}$ stands for the label space. Meta-learning performs adaptation to $\task$ through $\support$, and learns meta-knowledge through evaluation using $\query$. In few-shot classification, $\support$ and $\query$ share the same $N$ classes, but the samples of the classes are different from each other. If a class of $\support$ is composed of $K$ samples, the corresponding $\task$ is called $N$-way $K$-shot task and $|\support|$ becomes $NK$.

        MAML defines an initialization $\boldtheta$ as meta-knowledge and learns the initialization. Based on $\boldtheta$, MAML can complete adaptation to a new task $\task$ in only a small number of steps. Given $\{ \task \}_{i=1}^{B}$, the following processes are performed for each $\task$. First, a model is optimized on $\support$ in the inner-loop starting from $\boldtheta$ to get a task-specific parameter $\boldtheta_{i}$. Then, the adaptation proceeds according to the following equation:
        
        \begin{equation}
        \label{MAML:inner_loop}
            \boldtheta_{i} = \boldtheta - \alpha \nabla_{\boldtheta} \mathcal{L}_{\task}(\boldtheta;\support)
        \end{equation}

        \noindent where $\alpha$ is the learning rate in the inner-loop. Next, based on $\boldtheta_{i}$, $\mathcal{L}_{\task}(\boldtheta_{i};\query)$, i.e., loss for $\query$ is calculated. Finally, if all the processes are completed for every task in the mini-batch, $\boldtheta$ is updated by Eq.~(\ref{MAML:outer_loop}). 
        
        \begin{equation}
        \label{MAML:outer_loop}
            \boldtheta \gets \boldtheta - \eta \nabla_{\boldtheta} \sum_{i=1}^{B} \mathcal{L}_{\task}(\boldtheta_{i};\query)
        \end{equation}
        
        \noindent where $\eta$ is the learning rate in the outer-loop. In other words, how well the meta-knowledge is learned is evaluated and $\boldtheta$ is updated based on the evaluation. Therefore, the entire procedure results in learning well-generalized $\boldtheta$.

    \subsection{Contextual Gradient Scaling}
    \label{sec3.3:CGS}
        Gradient norm is one of the important indicators of how well a model is learning. So, its analysis is worthwhile. Let's take a look at the gradient norm of each layer in the inner-loop. In Figure~\ref{fig:gradient_norm_MAML}, we can find that not only the gradient norm of the classifier is much larger than that of each backbone layer, but also gradient norms of high-level layers of the backbone are smaller than those of the other layers. Due to this phenomenon, MAML tends to concentrate on decision boundary-oriented update and low-/mid-level features. Thus, MAML does not achieve effective adaptation in the inner-loop. There were a few naive approaches to solve this problem, e.g., freezing the classifier~\cite{BOIL} or clipping the gradients of the classifier. However, such approaches do not affect gradient norms of the backbone explicitly. 
        
        Thus, this paper proposes an explicit solution, CxGrad, that contextually scales gradient norms. CxGrad utilizes a sub-network $g_{\boldphi}$ and context parameters $\boldnu_{i}$. $g_{\boldphi}$ plays an important role in generating scaling factors and $\boldnu_{i}$ can be interpreted as embeddings of tasks. First, $g_{\boldphi}$ produces task-specific scaling factors using $\boldnu_{i}$. Then, in order to simplify the scaling procedure in the inner-loop, we use the BN property. 
        
        Before explaining the scaling process in detail, we need to briefly review the property of BN. While BN is scale-invariant in forward propagation, it is not in backward propagation. Let's take a look at this property mathematically. Following the notations of \cite{riemannian}, let the parameter vector of a convolution layer be $\boldv$, the scaled version of $\boldv$ be $\hat{\boldv} = a \boldv$, a vector of the input feature map be $\boldx$, and the covariance matrix be $\boldSigma$. Here, $a$ is a positive real number. Also, the output feature map is $\boldz = \boldv^{\intercal} \boldx$, and the scaled feature map is $\hat{\boldz} = \hat{\boldv}^{\intercal} \boldx$. Then, the outputs of BN layers before and after scaling the parameter vector by a scaling factor $a$ during forward propagation have the following relationship.
        
        \begin{equation}
        \begin{aligned}
        \mathrm{BN}(\hat{\boldv}^{\intercal} \boldx) &= \dfrac{\hat{\boldv}^{\intercal}(\boldx - \mathrm{E}[\boldx])}{\sqrt{\hat{\boldv}^{\intercal} \Sigma \hat{\boldv}}}
        = \dfrac{a \boldv^{\intercal}(\boldx - \mathrm{E}[\boldx])}{\sqrt{a^{2} \boldv^{\intercal} \Sigma \boldv}}\\[2ex]
        &= \dfrac{\boldv^{\intercal}(\boldx - \mathrm{E}[\boldx])}{\sqrt{\boldv^{\intercal} \Sigma \boldv}}
        = \mathrm{BN}(\boldv^{\intercal} \boldx)
        \end{aligned}
        \end{equation}
        
        \noindent However, in case of backward propagation, this scale-invariance does not hold. Instead, the gradient norm is scaled in proportion to the reciprocal of $a$.
        
        \begin{equation}
            \dfrac{\partial \mathrm{BN}(\hat{\boldz})}{\partial \hat{\boldv}} = \dfrac{1}{a} \dfrac{\partial \mathrm{BN}(\boldz)}{\partial \boldv}
            \label{eq:bn_backward}
        \end{equation}

        \noindent Note that since the BN property also holds even in Layer normalization~\cite{layernorm} and Group normalization~\cite{groupnorm}, the scaling process can be applied to any architectures containing the other normalization layers as well as BN layers. 
        
        Based on the BN property, gradient norms of the backbone are properly scaled using a scaling factor set. Let the scaling factor set of the backbone be $\mathcal{G} = \{\gamma^{(\ell)}_{i}\}^L_{\ell=1}$. Here, $\ell$ denotes an index of the backbone layer and $L$ denotes the number of backbone layers. Also, let the parameters of the backbone and classifier constituting a model be $\boldw_{i}, \boldpsi_{i}$, respectively. Then, CxGrad generates $\mathcal{G}$ from $g_{\boldphi}$ which is a sub-network parameterized with $\boldphi$. If a scaling factor $\gamma_{i}^{(\ell)}$ is always a positive number less than 1, the norm of the $\ell$-th backbone layer parameter $\|\boldw_{i}^{(\ell)}\|$ can increase without any upper bound because each gradient norm is scaled by the reciprocal of the scaling factor according to Eq.~(\ref{eq:bn_backward}). In order to avoid this, we $\ell_{2}$-normalize $\boldw_{i}^{(\ell)}$ to make $\|\boldw_{i}^{(\ell)}\|=1$ before scaling. Finally, the entire scaling procedure can be expressed as follows:
        
        \begin{algorithm}[t]
\fontsize{9}{11}\selectfont
\caption{Contextual Gradient Scaling (CxGrad)}

\makeatletter
\newcommand{\algmargin}{\the\ALG@thistlm}
\makeatother
\algnewcommand{\parState}[1]{\State%
  \parbox[t]{\dimexpr\linewidth-\algmargin}{\strut #1\strut}}

\textbf{Require: } $\alpha, \beta, \eta$: Learning rates\\
\textbf{Require: } $p(\mathcal{T})$: Distribution over tasks 

\begin{algorithmic}[1]
\State Randomly initialize $\boldtheta$ and $\boldphi$

\While{not done}
    \State Sample a mini-batch of tasks $\{ \task \}_{i=1}^{B}$ from $p(\mathcal{T})$
    
    \For{$i \in [B]$}
        \State $\boldtheta_{i} \gets \boldtheta$
        \State Reset context parameters: $\boldnu_{i} = \bm{0}$
        \parState{%
            Sample support and query sets ($\support, \query$) from $\task$}
        \State $\boldtheta_{i}^{'} = $\texttt{IGS}($\boldtheta_{i}, \boldnu_{i}$)
        \State Compute $\nabla_{\boldnu_{i}} \mathcal{L}(\boldtheta_{i}^{'};\support)$ using support sets.
        \parState {%
            Update context parameters: \\
            $\boldnu_{i} \gets \boldnu_{i} - \beta \nabla_{\boldnu_{i}} \mathcal{L}(\boldtheta_{i}^{'};\support)$}
        
        \For {the number of inner-loop steps}
            \State $\boldtheta_{i} \gets $\texttt{IGS}($\boldtheta_{i}, \boldnu_{i}$)
            \parState {%
                Compute $\nabla_{(\boldtheta_{i}, \boldnu_{i})} \mathcal{L}(\boldtheta_{i};\support)$ using support sets.}
            \parState {%
                Update both network and context parameters:\\
                $\boldtheta_{i} \gets \boldtheta_{i} - \alpha \nabla_{\boldtheta_{i}} \mathcal{L}(\boldtheta_{i};\support)$\\
                $\boldnu_{i} \gets \boldnu_{i} - \beta \nabla_{\boldnu_{i}} \mathcal{L}(\boldtheta_{i};\support)$}
        \EndFor
    
    \State Compute $\mathcal{L}(\boldtheta_{i};\query)$ using query sets.
    \EndFor
    
    \parState {
        Update the model parameters: \\
        $(\boldtheta, \boldphi) \gets (\boldtheta, \boldphi) - \eta \nabla_{(\boldtheta, \boldphi)}\sum_{i=1}^{B} \mathcal{L}(\boldtheta_{i};\query)$
        }
    
\EndWhile

\end{algorithmic}
\label{alg:CxGrad}
\end{algorithm}

\begin{algorithm}[t]
\fontsize{9}{11}\selectfont
\caption{Implicit Gradient Scaling (IGS)}

\makeatletter
\newcommand{\algmargin}{\the\ALG@thistlm}
\makeatother
\algnewcommand{\parState}[1]{\State%
  \parbox[t]{\dimexpr\linewidth-\algmargin}{\strut #1\strut}}

\textbf{Input: } Model parameters $\boldtheta_{i} = \{ \boldomega_{i}, \boldpsi_{i} \}$\\ \textbf{Input: } Context parameters $\boldnu_{i}$

\begin{algorithmic}[1]

\parState {%
Generate a set of scaling factors: \\ $\{\gamma^{(\ell)}_{i}\}^L_{\ell=1} = g_{\boldphi}(\boldnu_{i})$}
\parState {%
Scales the parameters of the backbone: \\ $\boldomega_{i} \gets \bigg\{ \gamma^{(\ell)}_{i} \boldomega^{(\ell)}_{i} / \lVert \boldomega^{(\ell)}_{i} \rVert \bigg\}^{L}_{\ell=1}$}
\State \textbf{Return} $\boldtheta_{i}$

\end{algorithmic}
\label{alg:IGS}
\end{algorithm}
        
        \begin{equation}
            \boldw_{i} \gets \{\gamma_{i}^{(\ell)} \boldw_{i}^{(\ell)} / \|\boldw_{i}^{(\ell)}\|\}_{\ell=1}^{L}
        \label{eq:weight_scaling}
        \end{equation}
        
        \noindent Owing to this inherent nature of BN, during forward propagation, the model outputs do not change even if the scaling process is applied. In other words, the loss doesn't change. However, during backward propagation, gradient norms are scaled by the reciprocal of $\gamma^{(\ell)}_{i}$. Putting these two cases together, the scaling factor for the $\ell$-th layer can be defined by $\|\boldw_{i}^{(\ell)}\| / \gamma_{i}^{(\ell)}$.
        
        Next, let's look into the generation process of $\mathcal{G}$. $g_{\boldphi}$ that generates $\mathcal{G}$ is a two-layer MLP with ReLU and softplus. ReLU is located between the layers, and softplus is located at the end of the sub-network. The reason why we adopt softplus is that it also can output a positive number greater than 1. So the gradient norms also can be scaled in a decreasing direction. Besides, since softplus can be differentiated like the other layers in $g_{\boldphi}$, the model can learn $\boldphi$ and the context parameters $\boldnu_{i}$, which are the input to $g_{\boldphi}$, in an end-to-end manner.
        Meanwhile, the context parameters $\boldnu_{i}$ serve as a task embedding. If the learning rate for $\boldnu_{i}$ is $\beta$, $\boldnu_{i}$ is updated as follows:
        
        \begin{equation}
            \boldnu_{i} \gets \boldnu_{i} - \beta \nabla_{\boldnu_{i}} \mathcal{L}(\boldtheta_{i};\support)
        \label{eq:nu_update}
        \end{equation}
        
        \noindent We omit the task notation $\task$ from the task-specific loss function $\mathcal{L}_{\task}$ because we only focus on classification tasks in this paper. Learning via Eq.~(\ref{eq:nu_update}) is plausible because $g_{\boldphi}$ consists of differentiable layers and both $g_{\boldphi}$ and $\boldnu_{i}$ are included in the computation graph. Finally, $\mathcal{G}$ is generated with $\boldnu_{i}$ as an input to $g_{\boldphi}$.
        
        \begin{equation}
            \mathcal{G} = \{\gamma^{(\ell)}_{i}\}^L_{\ell=1} = g_{\boldphi}(\boldnu_{i})
        \end{equation}
        
        \noindent $\boldnu_{i}$ should only embed the representation of the $i$-th task. So, before adapting to the next task $\mathcal{T}_{i+1}$, $\boldnu_{i}$ must be reset to a zero vector as in \cite{ModGrad, CAVIA}.
        
        Some may ask how about scaling gradient norms when updating $\boldw_{i}$. However, to update $\boldnu_{i}$, forward and backward propagation must be performed twice each. The first forward and backward propagation indicate the process of including $\boldnu_{i}$ in the computation graph (Eq.~(\ref{eq:weight_scaling})) and the process of updating $\boldnu_{i}$ (Eq.~(\ref{eq:nu_update})), respectively. Then, $\mathcal{G}$ is generated with the updated $\boldnu_{i}$ used in the second forward propagation. Finally, the gradients calculated by the second backward propagation are scaled, and then $\boldtheta_{i}$ is updated. On the other hand, the above process can be simplified by the BN property. Suppose we update $\boldnu_{i}$ before the first inner-loop step. Then, gradient norms can be scaled through only one pair of forward and backward propagation during the subsequent adaptation process. Furthermore, $\nabla_{\boldnu_{i}} \mathcal{L}_{\task}(\boldtheta_{i};\support)$ can be computed simultaneously with $\nabla_{\boldtheta_{i}} \mathcal{L}_{\task}(\boldtheta_{i};\support)$ without any extra process. For details, please refer to Algorithm~\ref{alg:CxGrad} and Algorithm~\ref{alg:IGS}.
        
\section{Experiments}
\label{sec:experiments}
    Various experiments show how efficient the proposed method is in few-shot classification. First, the proposed method is compared with several state-of-the-art (SOTA) algorithms on benchmark datasets, and then its quantitative evaluation in a cross-domain environment is given. In addition, we compare feature similarity at each backbone layer before and after adaptation, and visualize the effect of the proposed method using t-SNE~\cite{t-SNE}. Finally, various ablation studies are provided. In all the experiments below, four convolutional blocks, proposed by \cite{MatchingNet}, are used as the backbone architecture. 
    For more implementation details, please refer to the supplementary material. The codes are available at \url{https://github.com/shlee625/CxGrad}.

    \subsection{Few-Shot Classification}
    \label{exp:benchmark}

\begin{table*}
\setlength{\abovecaptionskip}{0pt}
\begin{center}
\begin{tabularx}{\textwidth}{>{\centering\arraybackslash}X>{\centering\arraybackslash}X
>{\centering\arraybackslash}X>{\centering\arraybackslash}X>{\centering\arraybackslash}X}
\toprule 
\multirow{2}{*}{\textbf{Method}} & \multicolumn{2}{c}{\textbf{miniImageNet}} & \multicolumn{2}{c}{\textbf{tieredImageNet}} \\
& \textbf{1-shot} & \textbf{5-shot} & \textbf{1-shot} & \textbf{5-shot} \\
\midrule
MatchingNet~\cite{MatchingNet}         & $43.44 \pm 0.77\%$          & $55.31 \pm 0.73\%$      & --                  & --\\
RelationNet~\cite{RelationNet}         & $50.44 \pm 0.82\%$          & $65.32 \pm 0.70\%$      & --                  & --\\
ProtoNet~\cite{ProtoNet}               & $49.42 \pm 0.78\%$          & $68.20 \pm 0.66\%$      & --                  & --\\
MAML~\cite{MAML}                       & $48.70 \pm 1.84\%$          & $63.11 \pm 0.92\%$      & $49.06 \pm 0.50\%$  & $67.48 \pm 0.47\%$\\
CAVIA (512)~\cite{CAVIA}               & $51.82 \pm 0.65\%$          & $65.85 \pm 0.55\%$      & --                  & --\\
BOIL~\cite{BOIL}                       & $49.61 \pm 0.16\%$          & $66.45 \pm 0.37\%$      & $48.58 \pm 0.27\%$  & $69.37 \pm 0.12\%$\\
MAML++~\cite{MAML++}                   & $52.15 \pm 0.26\%$          & $68.32 \pm 0.44\%$      & --                  & --\\
ALFA~\cite{ALFA}                       & $50.58 \pm 0.51\%$          & $69.12 \pm 0.47\%$      & $53.16 \pm 0.49\%$  & $70.54 \pm 0.46\%$\\
ModGrad~\cite{ModGrad}                 & \vcsb{53.20 \pm 0.86\%}     & $69.17 \pm 0.69\%$      & --                  & --\\
L2F~\cite{L2F}                         & $52.10 \pm 0.50\%$          & $69.38 \pm 0.46\%$      & $54.40 \pm 0.50\%$  & $73.34 \pm 0.44\%$\\
CxGrad (Ours)                          & $51.80 \pm 0.46\%$          & \vcsb{69.82\pm0.42\%} & \vcsb{55.55\pm0.46\%} & \vcsb{73.55\pm0.41\%}\\
\bottomrule
\end{tabularx}
\end{center}
\caption{Test accuracy on 5-way miniImageNet and tieredImageNet classification. CxGrad outperforms the others in 5-shot on both datasets and in 1-shot on tieredImageNet. CxGrad shows comparable result even in 1-shot on miniImageNet.}
\label{table:benchmark}
\end{table*}
        Two famous benchmark datasets, i.e., miniImageNet~\cite{MatchingNet} and tieredImageNet~\cite{tieredImageNet}, are used for few-shot classification. MiniImageNet composed of a total of 60,000 images was obtained by extracting 100 classes from ImageNet~\cite{ImageNet}. Each class contains 600 images of 84$\times$84. Here, 100 classes are again divided into three folds, i.e., 64, 16, and 20 classes which are used for meta-training, meta-validation, and meta-testing, respectively. The classes between folds do not overlap each other. TieredImageNet is a larger dataset than miniImageNet. Similar to miniImageNet, tieredImageNet is created by extracting 608 classes from ImageNet, and consists of a total of 779,165 images. It has the same spatial resolution as miniImageNet, but its classes are grouped into 34 upper categories unlike miniImageNet. Then, 20, 6, and 8 categories are used for meta-training, meta-validation, and meta-testing, respectively, and these categories do not overlap each other, either. Therefore, the datasets are suitable for evaluating the generalization ability of each algorithm.
        
        Table~\ref{table:benchmark} compares the proposed method with several meta-learning algorithms in 5-way 1-shot and 5-shot classification on miniImageNet and tieredImageNet. We can find that the proposed method noticeably outperforms MAML by 3.1\% to  6.49\% in all configurations. In particular, in the case of 5-shot on miniImageNet, it is noteworthy that the proposed method shows 0.53\% better than the SOTA algorithm. On the other hand, in the case of 1-shot on miniImageNet, the proposed method shows slightly lower than the SOTA algorithm. We think that since 1-shot has fewer samples per class than 5-shot, context parameters for embedding the tasks do not obtain sufficient information during meta-training. However, this is not the case with 1-shot on tieredImageNet. This is because despite the same 1-shot, sufficient meta-knowledge is learned thanks to the larger amount of iterations and datasets.

    \subsection{Cross-Domain Few-Shot Classification}
    \label{exp:cross-domain}
        To further analyze the effectiveness of our adaptation strategy, we examined few-shot classification in cross-domain environments. Unlike Section~\ref{exp:benchmark}, two datasets not used in meta-training were employed here for the tasks in meta-testing. Thus, it is necessary to learn a more optimized representation in the inner-loop for a task on the new dataset. In other words, the cross-domain experiment can better examine the efficacy of representation change. For this experiment, 5-way 5-shot was employed in the same architecture as Section~\ref{exp:benchmark}. After meta-knowledge is learned on miniImageNet, a meta-test was conducted by transferring the meta knowledge to the tasks on a different domain dataset.
        
        The datasets for the meta-test are CUB-200-2011~\cite{CUB} and CIFAR-FS~\cite{CIFAR-FS}. CUB has 200 fine-grained classes and consists of a total of 11,788 images. CUB is further divided into 100 meta-train classes, 50 meta-validation classes, and 50 meta-test classes. CIFAR-FS is a dataset based on CIFAR100~\cite{CIFAR100}. It has 100 classes, and each class consists of 600 images, for a total of 60,000 images. CIFAR-FS is again divided into 64 meta-train classes, 16 meta-validation classes, and 20 meta-test classes. Each image is resized to 84$\times$84.

\begin{table}[t]
\setlength{\abovecaptionskip}{0pt}
\setlength{\belowcaptionskip}{-5pt}
\begin{center}
\begin{tabular}{@{}ccc@{}}
\toprule 
\multirow{2}{*}{\textbf{Method}} & \multicolumn{2}{c}{\textbf{miniImageNet}} \\
& \textbf{$\to$ CUB} & \textbf{$\to$ CIFAR-FS} \\
\midrule
MAML          & $52.70 \pm 0.32\%$ & $55.82 \pm 0.50\%$\\
ALFA          & $58.35 \pm 0.25\%$ & $59.76 \pm 0.49\%$\\
BOIL          & $60.84 \pm 0.45\%$ & $63.28 \pm 0.45\%$\\
L2F           & $60.89 \pm 0.22\%$ & $63.73 \pm 0.48\%$\\
CxGrad (Ours) & \vcsb{63.92 \pm 0.44\%} & \vcsb{64.85 \pm 0.44\%}\\
\bottomrule
\end{tabular}
\end{center}
\caption{Test accuracy on 5-way 5-shot cross-domain few-shot classification on two meta-test datasets.}
\label{table:cross_domain}
\end{table}
        
        Table~\ref{table:cross_domain} compares CxGrad with several optimization-based algorithms. Note that the result of BOIL~\cite{BOIL} is reproduced and the others are from \cite{ALFA} as they are. As argued in \cite{BOIL}, if meta-training and meta-testing employ datasets in different domains respectively, representation change becomes further important as domain-specific knowledge is not meaningful any more. This is because domain-specific meta-knowledge does not cover the task distributions of the other domains. Here, domain-specific part in meta-knowledge is learned by samples from $p(\mathcal{T})$ of source domain, i.e., the domain of the meta-training dataset.  What can fill this knowledge gap between the domains is the representation change. CxGrad can alleviate the knowledge gap between domains as much as possible because it focuses on learning task-specific features from the target domain tasks. Therefore, CxGrad overwhelms the previous algorithms on CUB and CIFAR-FS. In particular, MAML's performance are improved with a large margin of at least $9\%$.

    \begin{figure}[t]
\setlength{\abovecaptionskip}{0pt}
\setlength{\belowcaptionskip}{5pt}
\begin{center}
    \includegraphics[width=.8\linewidth]{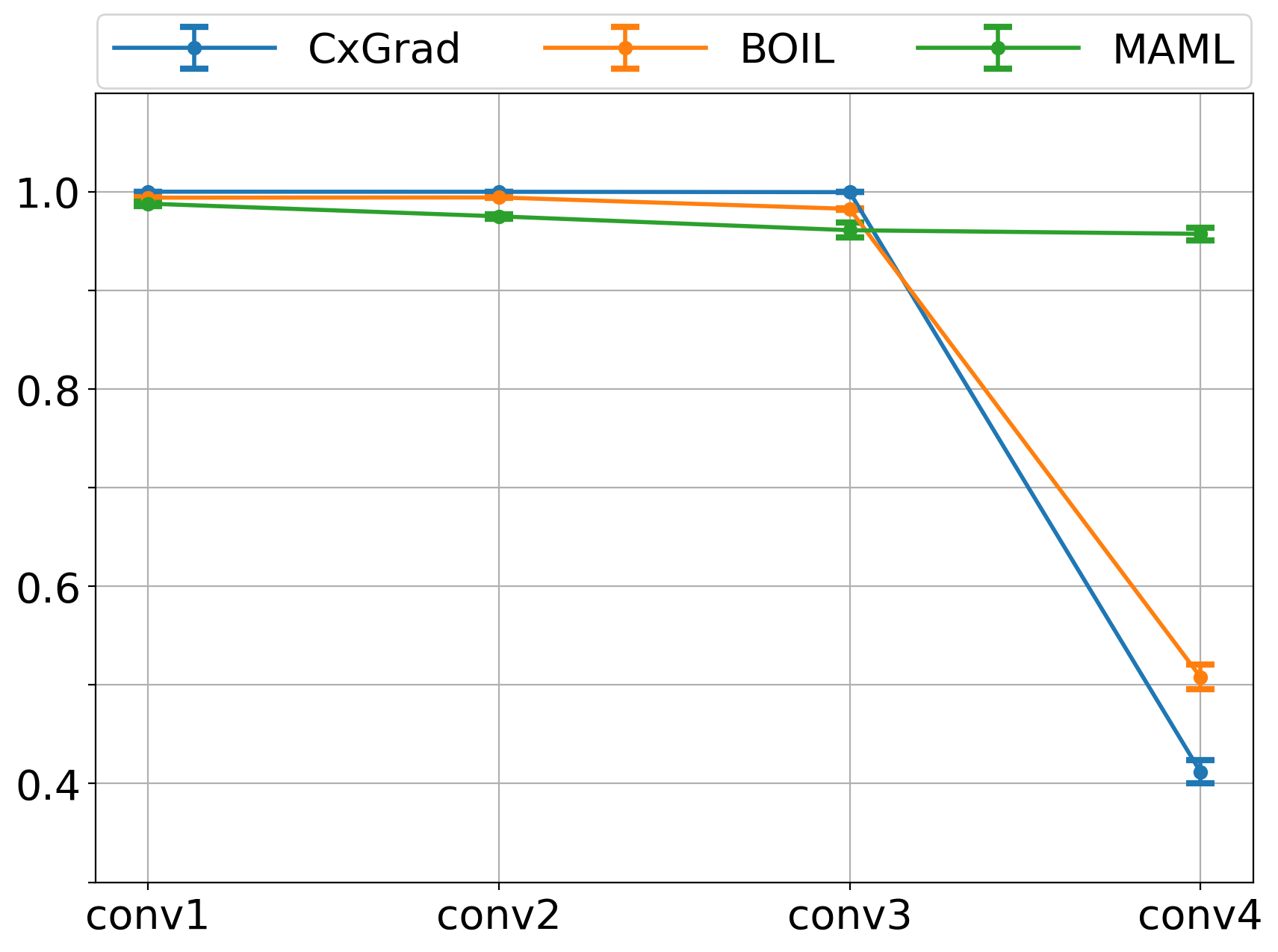}
\end{center}
\caption{CKA values of three optimization-based meta-learning algorithms. The x-axis means each layer in the backbone and the y-axis means the CKA value.}
\label{fig:feature_sim}
\end{figure}
    \begin{figure*}[t]
\begin{center}
    \begin{subfigure}[t]{.48\linewidth}
        \centering
        \begin{subfigure}[t]{.49\linewidth}
            \includegraphics[width=\linewidth]{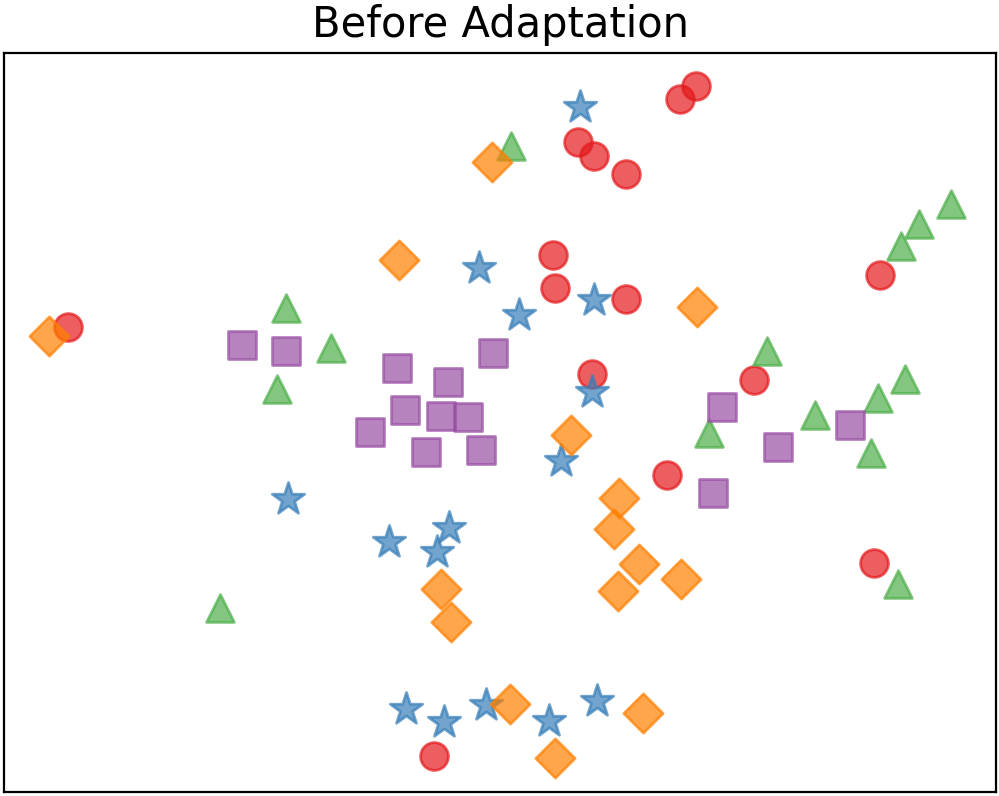}
        \end{subfigure}
        \begin{subfigure}[t]{.49\linewidth}
            \includegraphics[width=\linewidth]{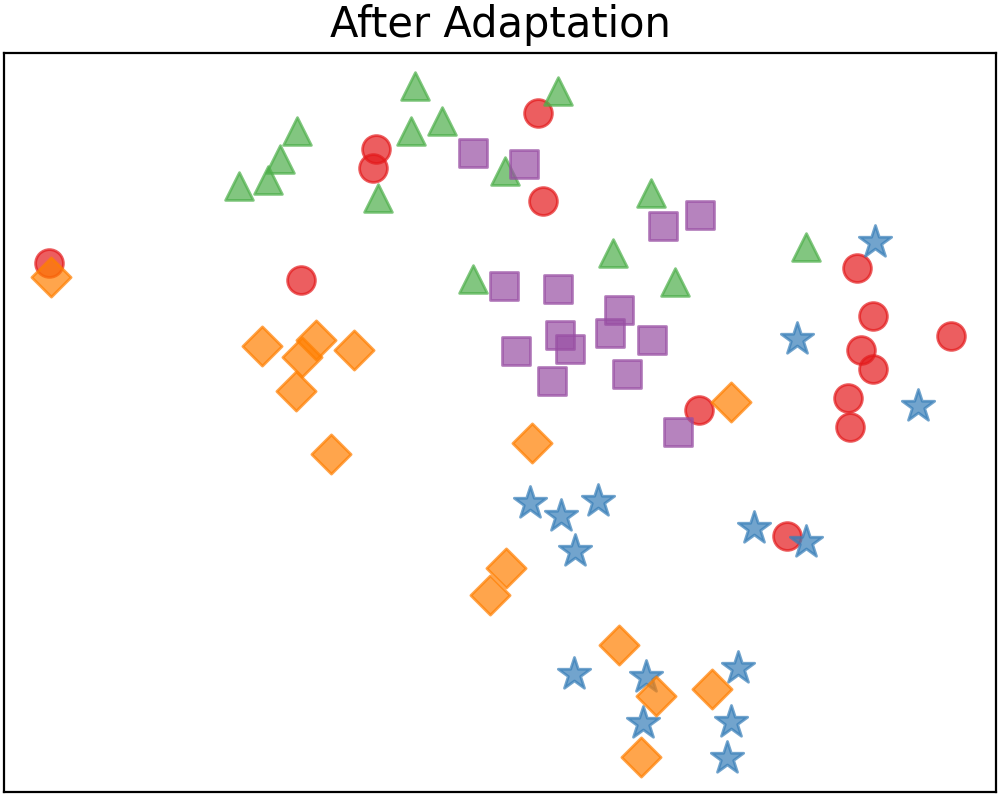}
        \end{subfigure}
    \end{subfigure}
    \hspace*{5mm}
    \begin{subfigure}[t]{.48\linewidth}
        \centering
        \begin{subfigure}[t]{.49\linewidth}
            \includegraphics[width=\linewidth]{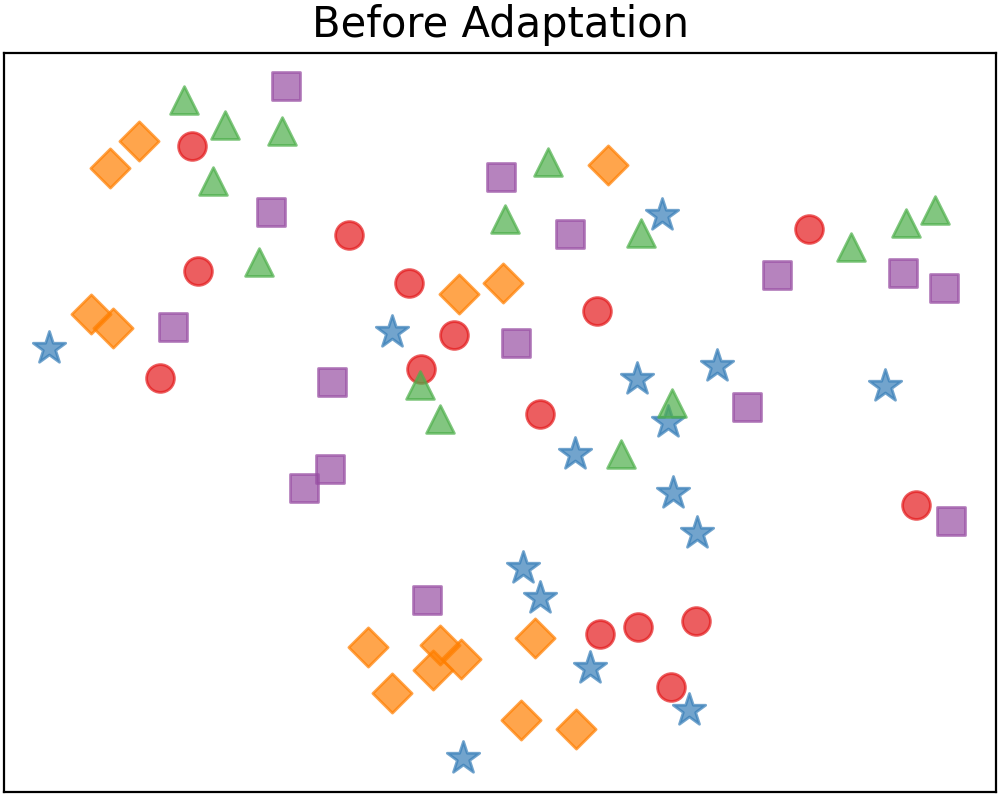}
        \end{subfigure}
        \begin{subfigure}[t]{.49\linewidth}
            \includegraphics[width=\linewidth]{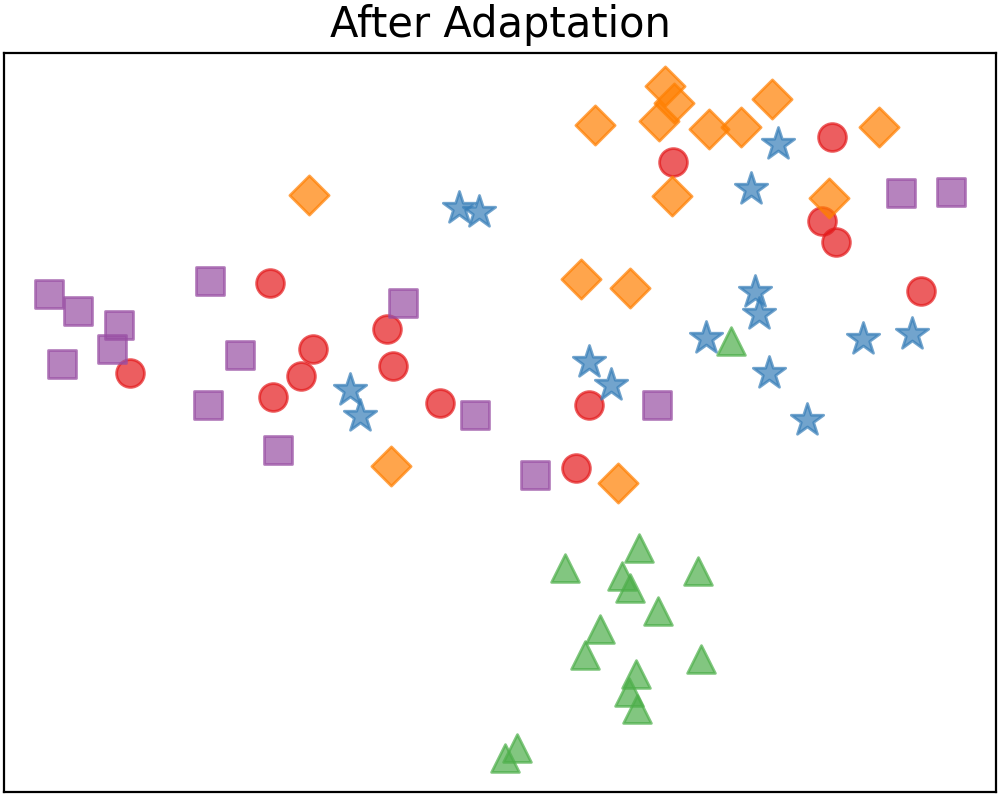}
        \end{subfigure}
    \end{subfigure}
    \end{center}
    
\begin{center}
    \begin{subfigure}[t]{.48\linewidth}
        \centering
        \begin{subfigure}[t]{.49\linewidth}
            \includegraphics[width=\linewidth]{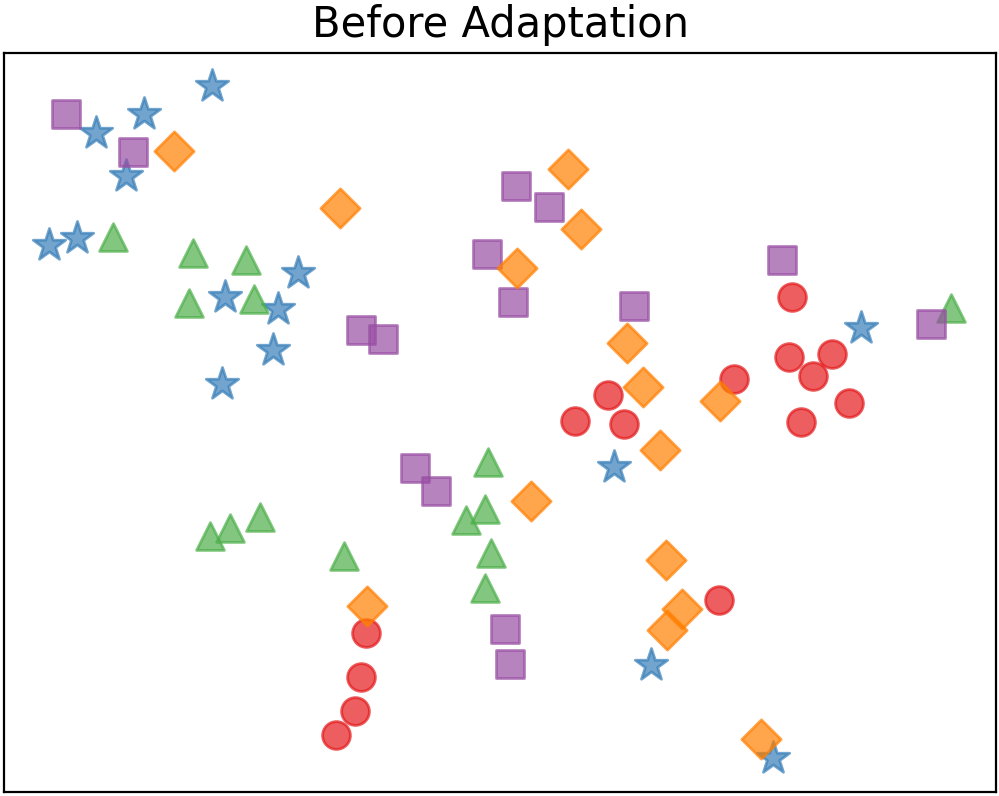}
        \end{subfigure}
        \begin{subfigure}[t]{.49\linewidth}
            \includegraphics[width=\linewidth]{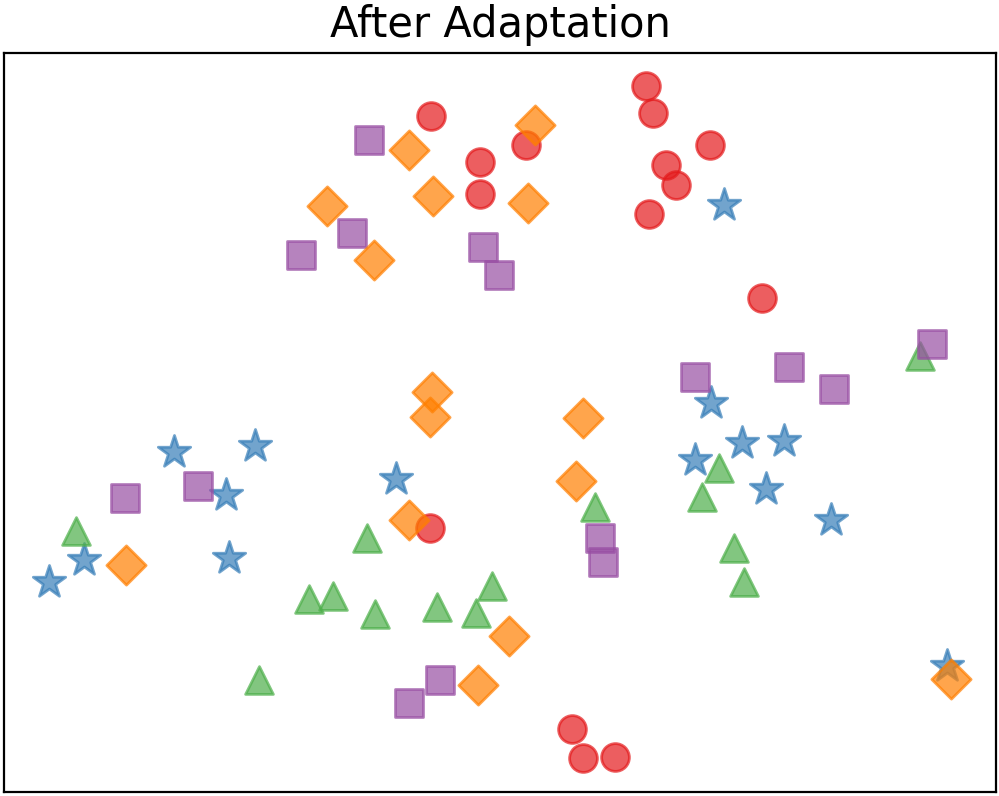}
        \end{subfigure}
        \caption{MAML}
    \end{subfigure}
    \hspace*{5mm}
    \begin{subfigure}[t]{.48\linewidth}
        \centering
        \begin{subfigure}[t]{.49\linewidth}
            \includegraphics[width=\linewidth]{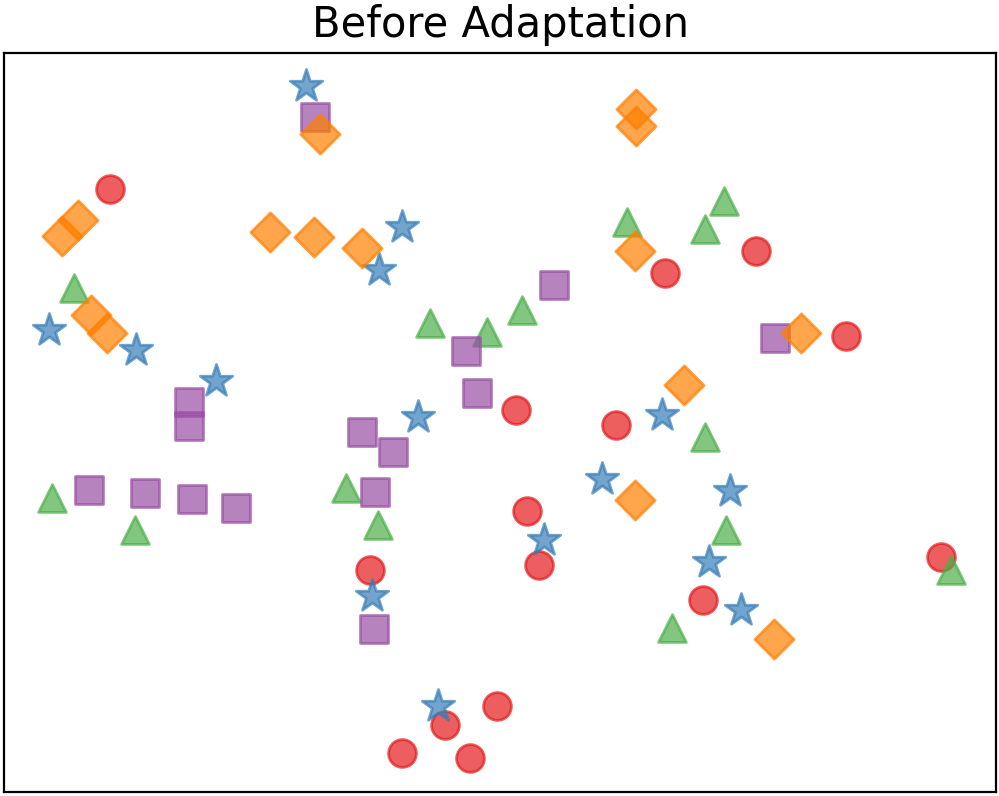}
        \end{subfigure}
        \begin{subfigure}[t]{.49\linewidth}
            \includegraphics[width=\linewidth]{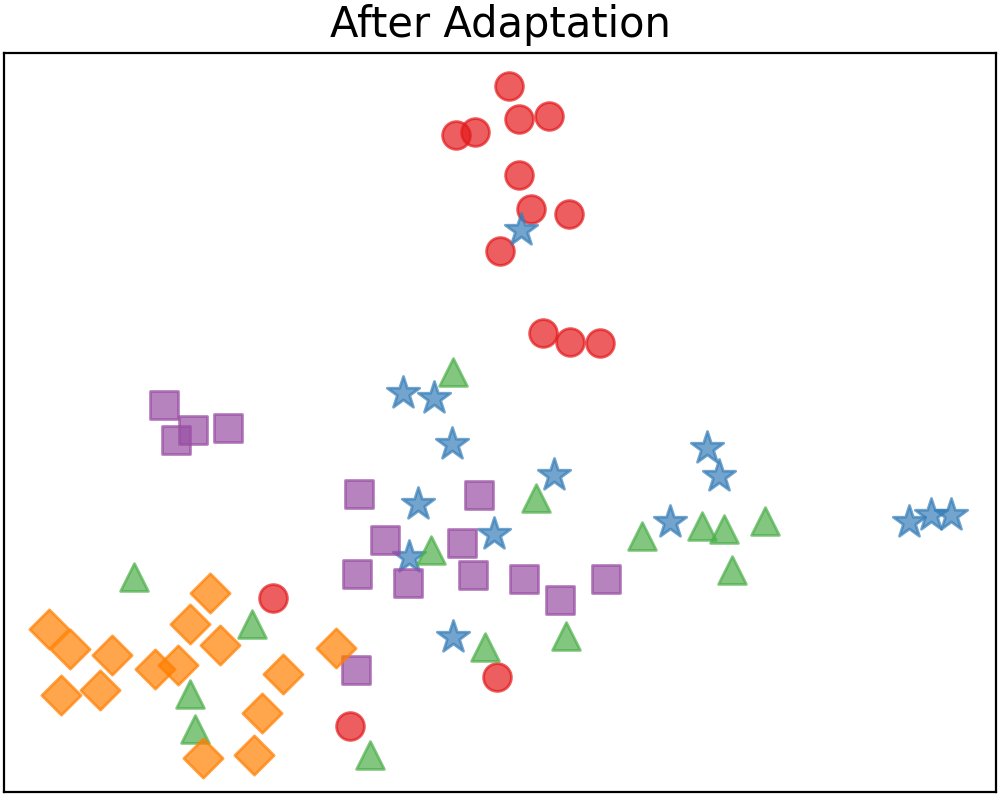}
        \end{subfigure}
        \caption{CxGrad}
    \end{subfigure}
\end{center}
\setlength{\abovecaptionskip}{0pt}
\caption{Visualization for latent feature representation using t-SNE. The first row is visualizations for meta-test on miniImageNet and the second row is on CUB. Each shape represents each class and the latent features are from the output of the last convolutional block of the backbone.}
\label{fig:embedding}
\end{figure*}
    
    \subsection{Representation Change}
    \label{exp:feature_similarity}
        In ANIL~\cite{ANIL} and BOIL~\cite{BOIL}, the layer-wise feature similarity before and after adaptation was measured and compared in terms of CKA~\cite{CKA}. CKA is one of the metrics to evaluate feature similarity. The larger the CKA value, the greater the similarity. By monitoring the CKA values, we can judge whether to reuse the existing representation or to use the representation optimized for a given task. We analyze MAML, BOIL, and the proposed method in the 5-way 5-shot on miniImageNet. In Figure~\ref{fig:feature_sim}, we plot the average and 95\% confidence interval of the CKA value among multiple tasks in meta-test phase. Note that we reproduced other algorithms and measured the CKA values.
        
        Looking at Figure~\ref{fig:feature_sim}, we can find that all three algorithms have CKA values close to 1 at the low- and mid-level layers. This indicates that they all learn task-generic low-level features as claimed in \cite{high-level_layer}. On the other hand, at the high-level layers, MAML still has CKA values near 1, but BOIL and CxGrad have low CKA values.  MAML remains the representation at the high-level layers, but BOIL and CxGrad utilize a more discriminative representation than MAML. Furthermore, CxGrad has a lower CKA value than BOIL. That is, CxGrad learns more task-specific features than BOIL as well as MAML by scaling gradient norms in the backbone.
    
    \subsection{Visualization of Embeddings}
    \label{exp:landscape}
        Figure~\ref{fig:embedding} visualizes how CxGrad embeds the tasks via t-SNE~\cite{t-SNE}, i.e., a nonlinear dimensionality reduction tool that maps high-dimensional data to low-dimensional space. First, we can observe that MAML does not perform effective representation change in both miniImageNet and CUB. Only a few classes that are properly clustered before adaptation make clusters after adaptation, while the other classes don't. On the other hand, CxGrad shows that samples dispersed before adaptation form a cluster after adaptation. It means that CxGrad indeed facilitates effective representation change in the inner-loop. Therefore, CxGrad can realize effective adaptation in the cross-domain datasets as well as the same domain datasets.

    \subsection{Ablation Study}
    \label{exp:ablation}
        \textbf{Update scheme for context parameters} A question may arise as to whether just updating context parameters for each step, i.e, $\boldnu_{i}$ is sufficient. We argue that it is useful for better performance to update context parameters through whole steps of the inner-loop. To prove this, the following ablation study compares two update schemes in 5-way 5-shot on miniImageNet: task-wise and step-wise. The task-wise method means updating the context parameters over all steps of the inner-loop for $\task$. On the other hand, the step-wise method means updating the context parameters for each separate step. In this case, we have to reset $\boldnu_{i}$ before moving on to the next step. Table~\ref{table:ablation_update_scheme} shows that task-wise update has better performance than step-wise update. This means that it is more advantageous to accumulate all knowledge of the parameters updated at each step.


\begin{table}[t]
\setlength{\abovecaptionskip}{0pt}
\begin{center}
\begin{tabular}{cc}
\toprule 
\textbf{Update scheme for $\boldnu_{i}$} & \textbf{Accuracy}\\
\midrule
step-wise & $68.83 \pm 0.43\%$\\
task-wise & \vcsb{69.82 \pm 0.42\%}\\
\bottomrule
\end{tabular}
\end{center}
\caption{Effect of context parameter update scheme. The accuracy results from 5-way 5-shot on miniImageNet.}
\label{table:ablation_update_scheme}
\end{table}
        
        \textbf{Effect of the number of steps} Now let's examine the effect of the number of inner-loop steps in adaptation. This experiment is also performed in 5-way 5-shot on miniImageNet. Table~\ref{table:ablation_num_steps} shows that although five steps shows the highest performance, the difference from a single step showing the lowest performance amounts to only 1\%. Therefore, the proposed method is relatively robust to the number of inner-loop steps.



\begin{table}[t]
\setlength{\abovecaptionskip}{0pt}
\begin{center}
\begin{tabular}{cccccc}
\toprule 
\textbf{Step} & \textbf{1} & \textbf{2} & \textbf{3} & \textbf{4} & \textbf{5}\\
\midrule
\textbf{Accuracy} & $69.12$ & $69.33$ & $69.16$ & $69.53$ & \vcsb{69.82}\\
\bottomrule
\end{tabular}
\end{center}
\caption{Ablation study on the number of inner-loop steps.}
\label{table:ablation_num_steps}
\end{table}

\begin{table}[t]
\setlength{\abovecaptionskip}{0pt}
\setlength{\belowcaptionskip}{-13pt}
\fontsize{9}{11}\selectfont
\begin{center}
\begin{tabular}{ccc}
\toprule 
& \textbf{Context parameter lr ($\beta$)} & \textbf{Accuracy}\\
\midrule
\multirow{3}{*}{1-shot} & $0.01$ & $51.56$ \\
& $0.1$ & $51.29$ \\
& $1.0$ & \vcsb{51.80} \\
\midrule
\multirow{3}{*}{5-shot} & $0.01$ & $69.28$\\
& $0.1$ & $69.34$\\
& $1.0$ & \vcsb{69.82}\\
\bottomrule
\end{tabular}
\end{center}
\caption{Ablation study on the learning rate for the context parameters.}
 \label{table:ablation_lr}
\end{table}

        
        \textbf{Inner-loop learning rate} The proposed method has another learning rate~($\beta$) related to context parameters. To examine the performance change according to the learning rate, we compare the models trained with three different learning rates. For convenience, the difference between the learning rates was fixed 10 times.
        This experiment was performed in 5-way 1-shot and 5-shot on miniImageNet. In Table~\ref{table:ablation_lr}), both cases show better performance when $\beta$ is 1.0. However, note that even if the learning rates differ by 100 times, the difference in performance does not exceed 1\%. Therefore, we can claim that the proposed method is meaningfully robust to hyperparameters.

\section{Conclusion}
\label{sec:conclusion}
    This paper argues that the ineffective adaptation of MAML results from the two reasons: 1) the gradients are concentrated on the classifier rather than the backbone and 2) the gradients are not concentrated on the high-level layer inside of the backbone. As a result, the performance of MAML is hindered in meta-testing. In particular, the performance degradation is more pronounced in a cross-domain environment. In order to solve this problem, we scale the gradient norms of the backbone by using the property of BN. Here, the scaling factors are generated from the context parameters which embed the task information. As a result, the gradients are scaled in a task-wise manner.
    Experimental results show that the proposed method accomplishes more effective adaptation. In particular, the knowledge gap caused by domain change in a cross-domain environment is solved by changing the representation through gradient scaling.

\section*{Acknowledgments}
\label{sec:acknowledgments}
    This research was supported by the MSIT(Ministry of Science and ICT), Korea, under the ITRC(Information Technology Research Center) support program (IITP-2021-0-02052) supervised by the IITP(Institute for Information \& Communications Technology Promotion) and IITP grant funded by MSIT [2020-0-01389, Artificial Intelligence Convergence Research Center(Inha University)].

{\small
\bibliographystyle{ieee_fullname}
\bibliography{egbib}
}

\renewcommand\thesection{\Alph{section}}
\setcounter{section}{0}
\clearpage
\newpage

\section{Experiment Settings}
\label{sup:experiment_detail}
    Our implementation and experiment settings are based on those of ALFA~\cite{ALFA}. In all the experiments, The network architecture is 4 convolutional blocks~\cite{MatchingNet} followed by a fully connected layer. Each block consists of a convolutional layer with 3$\times$3 kernel, a batch normalization layer, a ReLU activation function, and a 2$\times$2 max-pooling layer. The dimension of the context parameters and the dimension of the intermediate features in $g_{\boldphi}$ are both 100. The dimension of the outputs of $g_{\boldphi}$ is same as the number of convolution layers in the backbone. In $\query$, each class contains 15 samples. None of the data-augmentation methods are used in training. The batch size $B$ is 4 for 1-shot and 2 for 5-shot. We optimize the model in the inner-loop for 5 steps and set the learning rates $\alpha$, $\beta$, and $\eta$ to be 0.01, 1.0, 0.001, respectively. Adam~\cite{ADAM} is used as the meta-optimizer in outer-loop. Our model is trained for 50,000 iterations on miniImageNet and 125,000 iterations on tieredImageNet. An ensemble of models, whose ranks are top 5 in terms of meta-validation accuracy, is evaluated on 600 tasks from meta-test set. We run 3 independent runs with 3 different seeds and report the average results.

\section{Experimental Results on Bigger Backbone}
\label{sup:bigger_backbone}
    \begin{table}[t]
\setlength{\abovecaptionskip}{0pt}
\setlength{\belowcaptionskip}{-5pt}
\begin{center}
\begin{tabular}{@{}ccc@{}}
\toprule
\multirow{2}{*}{\textbf{Method}} & \multicolumn{2}{c}{\textbf{miniImageNet}} \\
& \textbf{1-shot} & \textbf{5-shot} \\
\midrule
MAML~\cite{MAML} & $58.37 \pm 0.49\%$ & $69.76 \pm 0.46\%$\\
BOIL~\cite{BOIL} & - & $71.30 \pm 0.28\%$\\
L2F~\cite{L2F}   & $59.71 \pm 0.49\%$ & $77.04 \pm 0.42\%$\\
ALFA~\cite{ALFA} & $59.74 \pm 0.49\%$ & \vcsb{77.96 \pm 0.41\%}\\
CxGrad (Ours)    & \vcsb{60.19\pm0.45\%} & $75.17\pm0.40\%$\\
\bottomrule
\end{tabular}
\end{center}
\caption{Test accuracy on 5-way miniImageNet classification.}
\label{table:resnet12_mini_imagenet}
\end{table}
    In this section, we provide additional experimental results on a deeper backbone, especially ResNet-12~\cite{resnet}. ResNet-12 consists of 4 residual blocks. Each residual block is composed of three convolutional blocks, each of which consists of a convolutional layer, a BN layer, and a ReLU activation function. A pointwise convolutional block is positioned at the skip connection for matching the number of channels between the residual inputs and the outputs. A 2$\times$2 max-pooling layer is at the end of each residual block. The number of channels begins with 64 and gets doubled by each residual block. Finally, we aggregate the spatial dimension of the final representation by a global average pooling layer and pass it to the classifier. For ResNet-12, we apply the scaling process for all the residual blocks. More specifically, in each residual block, we only scale the weights for the three convolutional blocks, not for the pointwise convolutional block. Table~\ref{table:resnet12_mini_imagenet} and Table~\ref{table:resnet12_tiered_imagenet} provide 5-way few-shot classification performance using ResNet-12 on miniImageNet and tieredImageNet, respectively.
    \begin{table}[t]
\setlength{\abovecaptionskip}{0pt}
\setlength{\belowcaptionskip}{-5pt}
\begin{center}
\begin{tabular}{@{}ccc@{}}
\toprule
\multirow{2}{*}{\textbf{Method}} & \multicolumn{2}{c}{\textbf{tieredImageNet}} \\
& \textbf{1-shot} & \textbf{5-shot} \\
\midrule
MAML~\cite{MAML} & $58.58 \pm 0.49\%$  & $71.24 \pm 0.43\%$ \\
L2F~\cite{L2F}   & $64.04 \pm 0.48\%$  & $81.13 \pm 0.39\%$ \\
ALFA~\cite{ALFA} & $64.62 \pm 0.49\%$  & $82.48 \pm 0.38\%$ \\
CxGrad (Ours)    & \vcsb{65.47\pm0.44\%} & \vcsb{82.52\pm0.35\%} \\
\bottomrule
\end{tabular}
\end{center}
\caption{Test accuracy on 5-way tieredImageNet classification.}
\label{table:resnet12_tiered_imagenet}
\end{table}

\section{Optimization Landscape}
\label{sup:loss_landscape}
    \begin{figure*}[t]
\begin{center}
    \begin{subfigure}[t]{.33\linewidth}
        \centering
        \includegraphics[width=1\linewidth]{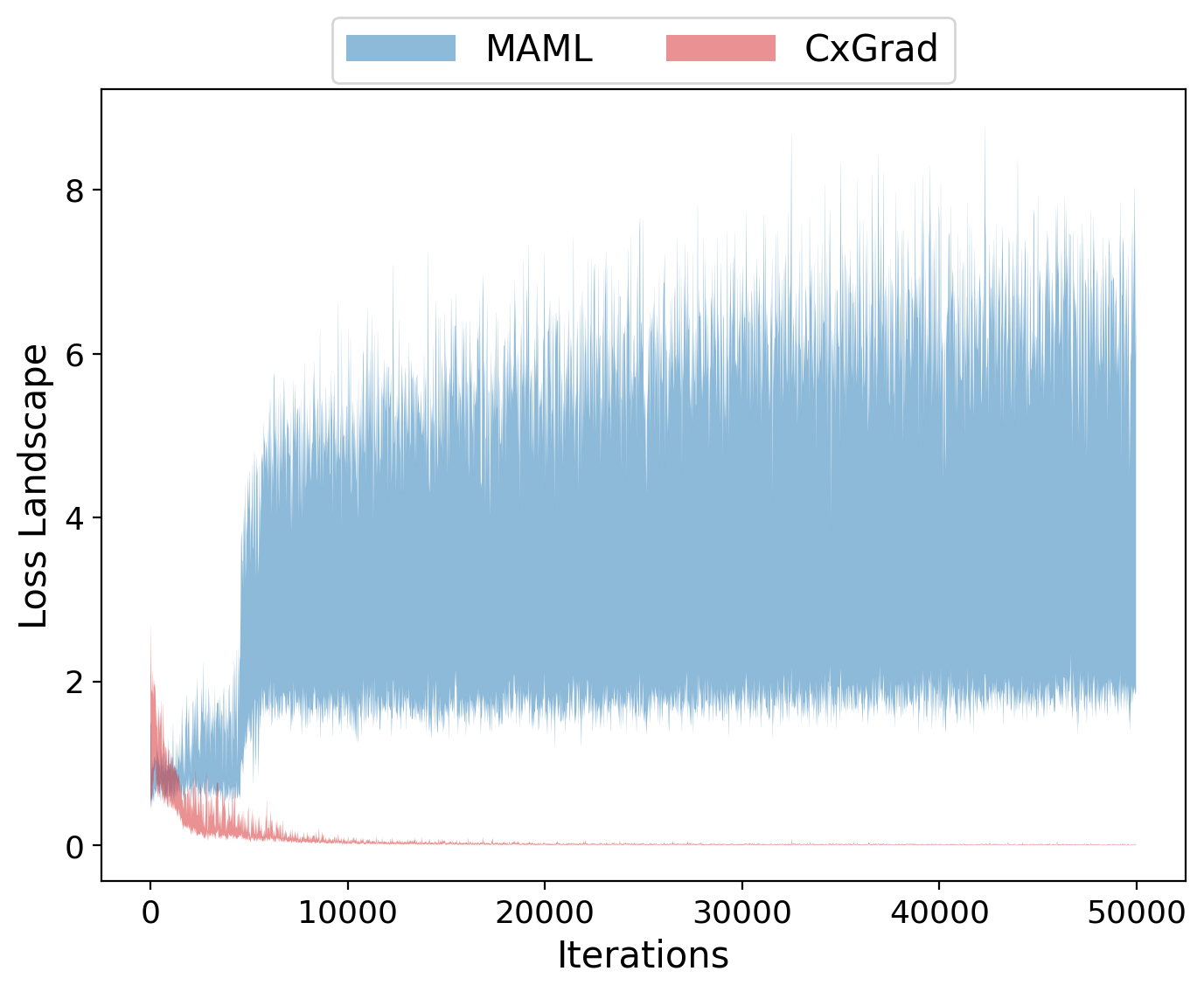}
        \caption{Loss landscape}
        \label{fig:landscape_loss}
    \end{subfigure}
    \begin{subfigure}[t]{.33\linewidth}
        \centering
        \includegraphics[width=1\linewidth]{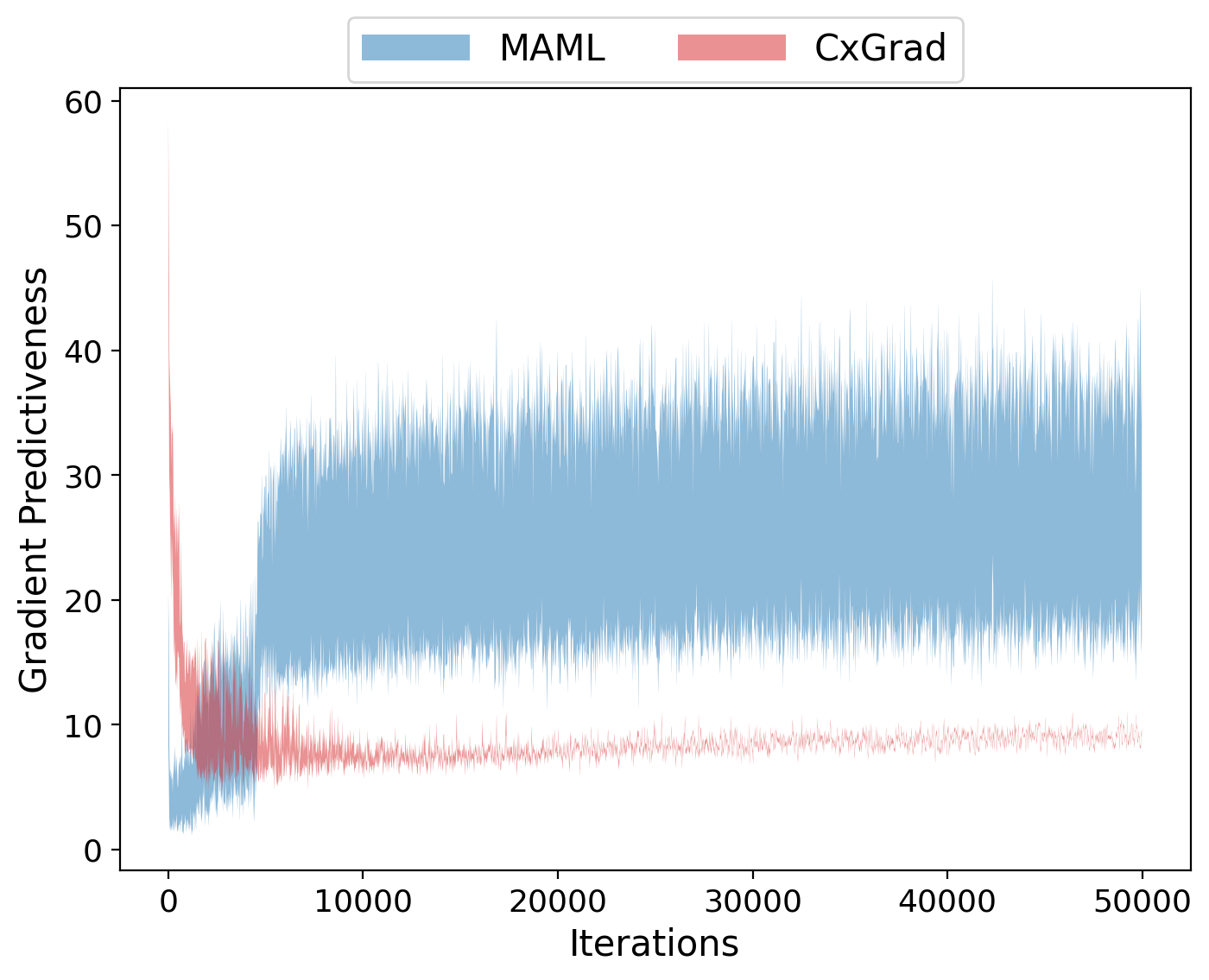}
        \caption{Gradient predictiveness}
        \label{fig:landscape_gradient}
    \end{subfigure}
    \begin{subfigure}[t]{.33\linewidth}
        \centering
        \includegraphics[width=1\linewidth]{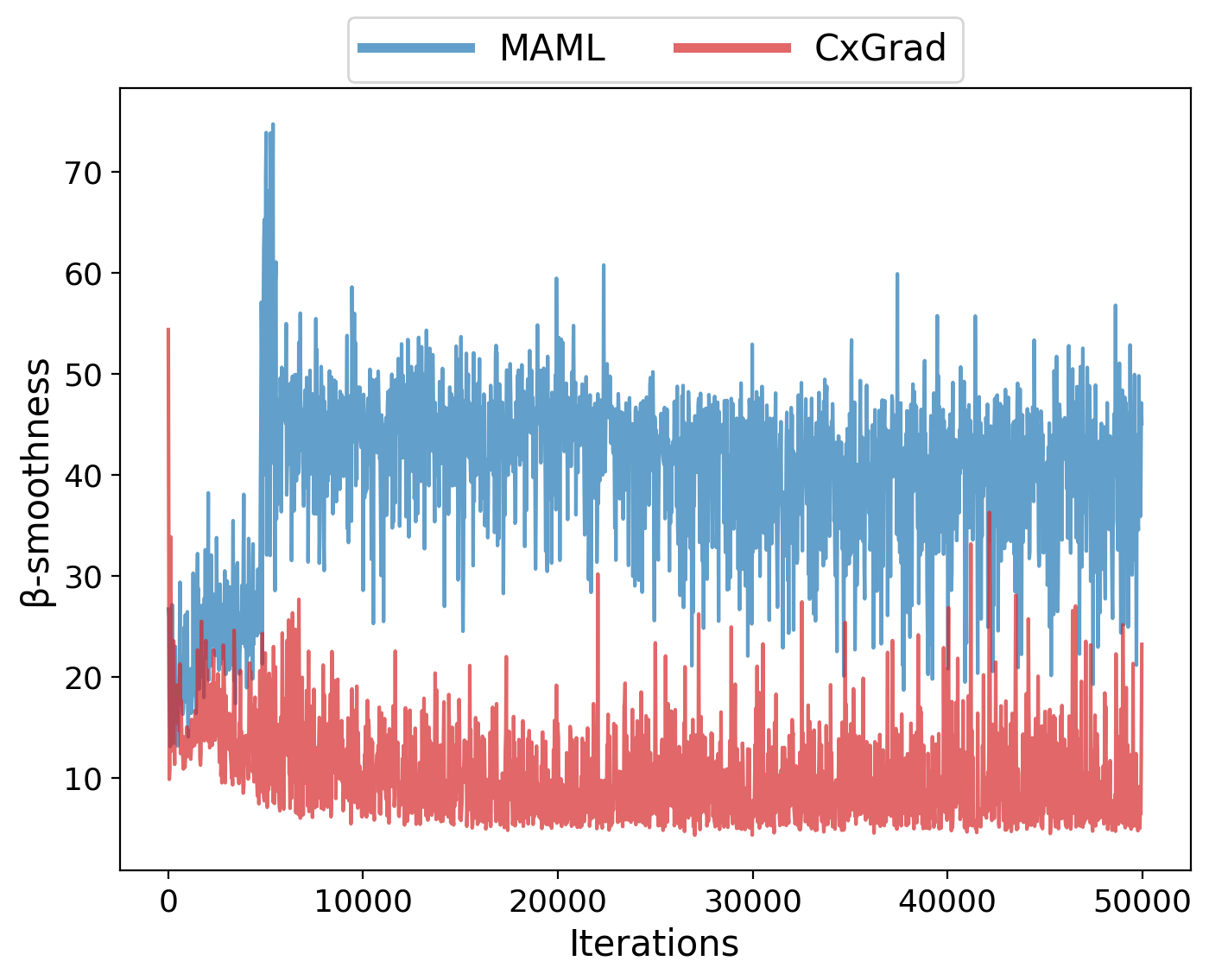}
        \caption{``effective'' $\beta$ smoothness}
        \label{fig:landscape_beta}
    \end{subfigure}
\end{center}
    \setlength{\abovecaptionskip}{0pt}
   \caption{Optimization landscape during adaptation. Like other experiments, we plot these results from 5-way 5-shot miniImageNet few-shot classification. In order to analyze the landscape, we update the model parameter from a certain inner-loop step with various range of learning rates. We refer to these updated parameters as \textit{points}. With these, we can analyze the local landscape from the certain step. (a) We measure the variation in loss calculated at the \textit{points}. (b) We measure the variation in $\ell_{2}$ distance between the gradients at the certain step and at each \textit{point}. (c) ``effective'' $\beta$ smoothness refers to the maximum $\ell_{2}$ difference of (b) over distance as we move to the corresponding \textit{point} from the certain step.}
\label{fig:landscape}
\end{figure*}
    Ioffe~\cite{batchnorm} argued that batch normalization helps the training by reducing \textit{internal covariate shift} (ICS). However, Santurkar~\cite{how_does_BN} found that the true reason is that batch normalization smooths the optimization landscape in training. In other words, batch normalization improves the Lipschitzness of both the loss and the gradients of a model. To prove this argument, he measured the variation in loss, gradient predictiveness, and ``effective'' $\beta$ smoothness in the vicinity of a certain point on the optimization landscape. For more details, please refer to Section 3 in \cite{how_does_BN}. Based on this study, Baik~\cite{L2F} analyzed the optimization landscape of MAML and his method L2F. Likewise, in this section, we also analyze how our method affects the optimization landscape in the inner-loop following \cite{L2F,how_does_BN}. In Figure~\ref{fig:landscape}, we plot these three measurements and explain the meaning of each one.
    
    In order to analyze the optimization landscape in the inner-loop, we have to observe the loss and the gradients in the vicinity of the model parameters adapted to $\task$. To accomplish this, we perform adaptation with a new learning rate set in $[0.5, 4] \times \alpha$. We set $\alpha = 0.01$ as before. Let the new learning rate set $\mathcal{A} = \{x|x=0.5 \alpha i,i\in[8]\}$, $j$-th learning rate of $\mathcal{A}$ be $\bar{\alpha}_{j}$, and $\boldtheta_{i,s}$ be the parameters of the adapted model to $\task$ at $s$-th step in the inner-loop. Then, we can compute several model parameters around $\boldtheta_{i,s}$ as below:
    
    \begin{equation}
        \boldtheta_{i,s}^{\bar{\alpha}_{j}} = \boldtheta_{i,s} - \bar{\alpha}_{j} \nabla_{\boldtheta_{i,s}} \mathcal{L}(\boldtheta_{i,s};\support)
    \end{equation}
    
    \noindent Next, let $B$ denote the batch size and $S$ denote the number of inner-loop steps. We plot the variation in the loss and the gradient predictiveness computed by Eq.~(\ref{eq:loss_landscape}) and Eq.~(\ref{eq:gradient_predictiveness}) in Figure~\ref{fig:landscape_loss} and Figure~\ref{fig:landscape_gradient}, respectively.
    
    \begin{equation}
    \label{eq:loss_landscape}
        \dfrac{1}{BS} \sum_{b=1}^{B} \sum_{s=1}^{S} \mathcal{L}(\boldtheta_{i, s}^{\bar{\alpha}_{j}}; \support)
    \end{equation}
    
    \begin{equation}
    \label{eq:gradient_predictiveness}
        \dfrac{1}{BS} \sum_{b=1}^{B} \sum_{s=1}^{S} \| h(\boldtheta_{i, s}^{\bar{\alpha}_{j}}) - h(\boldtheta_{i})\|
    \end{equation}
    
    \noindent where $h(\boldtheta_{i}) = \nabla_{\boldtheta_{i}}\mathcal{L}(\boldtheta_{i};\support)$. In Figure~\ref{fig:landscape_beta}, we calculate the values by Eq.~(\ref{eq:effective_beta}) and plot them. Here, $\bar{\alpha}_{j}^{*} = \argmax_{\bar{\alpha}_{j} \in \mathcal{A}} \| h(\boldtheta_{i, s}^{\bar{\alpha}_{j}^{*}}) - h(\boldtheta_{i})\|$.
    
    \begin{equation}
    \label{eq:effective_beta}
        \dfrac{1}{BS} \sum_{b=1}^{B} \sum_{s=1}^{S} \dfrac{\| h(\boldtheta_{i, s}^{\bar{\alpha}_{j}^{*}}) - h(\boldtheta_{i}) \|}{\| \bar{\alpha}_{j}^{*} h(\boldtheta_{i}) \|}
    \end{equation}

    Looking at Figure~\ref{fig:landscape_loss}, we can observe the loss landscape in the inner-loop as the training proceeds. In the case of MAML, the variation in loss becomes larger from approximately 5,000 iterations. It means that the loss landscape gets sharper in the vicinity of the points on it. On the contrary, in the case of CxGrad, the variation in loss is drastically reduced, implying that CxGrad efficiently and effectively smooths the loss landscape. In Figure~\ref{fig:landscape_gradient}, gradient predictiveness means how far the gradients around a certain point are from the gradient at the point. The farther the distance, the lower the stability. At the beginning of training, CxGrad is more unstable than MAML in terms of gradients because the sub-network $g_{\boldphi}$ doesn't learn sufficient knowledge from tasks to scale the gradients of the backbone in a task-wise manner. Nevertheless, CxGrad retains enough stability in no time. Lastly, in Figure~\ref{fig:landscape_beta}, CxGrad shows better Lipschitzness than MAML. It means that CxGrad also smooths the gradients besides the loss. As a result, CxGrad improves both the convergence speed and the performance of the model.

\end{document}